\documentclass{article}
\pdfoutput=1

\usepackage[final]{neurips_2019}

\usepackage[utf8]{inputenc} 
\usepackage[T1]{fontenc}    
\usepackage{hyperref}       
\usepackage{url}            
\usepackage{booktabs}       
\usepackage{amsfonts}       
\usepackage{nicefrac}       
\usepackage{microtype}      
\usepackage{xcolor}

\usepackage{amsmath}
\usepackage{amsthm}
\usepackage{multirow}
\usepackage{graphicx}
\usepackage{float}
\usepackage{subcaption}

\DeclareMathOperator*{\argmin}{arg\,min}

\newtheorem{theorem}{Theorem}

\title{Getting Topology and Point Cloud Generation to Mesh}

\author{Austin Dill\thanks{equal contribution}, Chun-Liang Li$^\star{}$, Songwei Ge, Eunsu Kang \\
Carnegie Mellon University\\
Pittsburgh, PA, United States \\
\{abdill, chunlial, songweig, eunsuk\}@andrew.cmu.edu\\
\newline
\newline
}

\begin{document}

\maketitle

\begin{abstract}

In this work, we explore the idea that effective generative models for point clouds under the autoencoding framework must acknowledge the relationship between a continuous surface, a discretized mesh, and a set of points sampled from the surface. This view motivates a generative model that works by progressively deforming a uniform sphere until it approximates the goal point cloud. We review the underlying concepts leading to this conclusion from computer graphics and topology in differential geometry, and model the generation process as deformation via deep neural network parameterization.
Finally, we show that this view of the problem produces a model that can generate quality meshes efficiently. 
    
\end{abstract}

\section{Introduction}

The proliferation of 3D point cloud data from LiDAR sensors for self-driving cars and commercial sensors like Microsoft's Kinect has led to an increase in the interest for machine learning techniques to interpret 3D data. Just as generative models for text and images gained traction following the success of classification models for those domains, the success of neural networks on point clouds~\citep{qi2017pointnet,qi2017pointnet++,zaheer2017deep} has led to a growth in research into point cloud generative models~\citep{yang2018foldingnet,li2018point,groueix2018atlasnet,yang2019pointflow,groueix20183d,li2019lbs}. 

In this paper, we study point cloud autoencoders, which are building components of a full point cloud generative model~\citep{li2018point,yang2019pointflow}. The decoder itself is a generative model which samples from a corresponding 3-dimensional distribution of the input point cloud.
If we naively train a neural network to generate point clouds, we will end up with an extremely under-constrained problem. This will result in low reconstruction loss but a non-uniform distribution of points on the surface of the object and it will require an additional meshing process. 

In order to remedy this, we draw inspiration from computer graphics and topology by learning a generative model by way of \textit{deformation}. We begin with a goal object we hope to reconstruct and an initial point cloud. At each step, we will learn a slight modification of the initial point cloud so the output will be close to the goal point cloud. As the underlying object that generated the point cloud sample is a solid object, we will constrain our model to learn deformations that obey the rules that govern valid topological and mesh deformations. As we will see, this will limit our network to \textit{invertible deformations}. Incidentally, by focusing on this type of deformation, we can frequently use a mesh from the initial point cloud as a template for our final object, simplifying the meshing pipeline. 

\section{Background}

 Traditional computer graphics approaches such as ball-pivoting, marching cubes, and Poisson reconstruction work by algorithmically building a mesh to fit a fixed set of points. 

\cite{Ezuz_2019} connect this generation of meshes to the problem of matching points on corresponding meshes. Their problem then becomes finding a plausible, smooth, and accurate map between triangular meshes. As investigated in their work, this requires a harmonic and reversible mapping between meshes that can be found by optimizing a loss function for individual pairs of meshes with a true underlying correspondence. This approach has the benefit of allowing the transference of texture and other mesh properties from the source mesh to the target mesh. 

The closest work to ours in the deep learning community is that of FoldingNet~\citep{yang2018foldingnet}. Their model learns a series of deformations (or folds) from an initial fixed 2D grid of points to a final object. It seems intuitive that starting from a 3D surface could lead to an easier learning problem, just as it is simpler to mold clay than to fold origami. This idea is generalized with AtlasNet where multiple 2D grids are used with multiple generators. In addition they explore using a sphere to sample points, but neglect the theoretical advantages connecting this approach to topology and the concept of deformations ~\citep{groueix2018atlasnet}. 

While our work focuses on the generation of point clouds, a similar vein of work has been explored in the mesh reconstruction field by works like \cite{Wang_2018} and \cite{Kanazawa_2018}. These methods both proceed by deforming an initial mesh (given a priori or learned respectively) into a final shape. These methods employ a graph-based method that does not allow for sampling an arbitrary number of points. 

\section{Theoretical Justification}

We must examine the complexity of transforming various initial distributions to the type of surfaces encoded with point clouds to justify our claim that our model is a more natural generative model for point clouds. To simplify matters, we will limit our discussion to maps from a $3$-ball and a 3-dimensional isotropic Gaussian to a $2$-sphere. 

Point clouds are typically sampled from the surfaces of real-world objects or realistic meshes. Intuitively, this means that the majority of points are located on the boundary of objects and not on the interior. If we hope to perfectly capture the surface of objects in point clouds with a continuous, invertible map as has become common practice in many generative models, we must consider the topology of our initial shape ~\citep{rezende2015variational, grathwohl2018ffjord, behrmann2018invertible}. 

\begin{theorem}
There is no continuous invertible map between the 3-ball and the 2-sphere that respects the boundary. 
\end{theorem}

\vspace{-1.5em}
\begin{proof}
This follows from Brouwer's fixed point theorem. 
\end{proof}

\begin{theorem}
There is no continuous invertible map between $\mathbb{R}^3$ and the 2-sphere that respects the boundary. 
\end{theorem}

\vspace{-1.5em}
\begin{proof}
This follows from the relationship between Hausdorf spaces and compact subspaces. 
\end{proof}

These results show that if we wish to learn a transformation that is continuous, invertible, and achieves no error on the boundary, we must choose an initial point cloud that is topologically close to our goal point cloud. Otherwise, our efforts will be thwarted by the underlying topology. For these reasons we choose to start from a hollow sphere of points with radius 1 as our initial shape. We believe that this is the most topologically similar structure that is simple to sample from. As we will see in section \ref{loss_function_section}, this decision gives us additional advantages. 

\section{Method}

\subsection{Architecture}

The architecture for our network is built on the idea of repeated deformations to an initial point cloud based on the encoding generated by a Deep Set model ~\citep{zaheer2017deep}. This model takes inspiration from FoldingNet with its series of "folds" replaced by deformations and its graph-based encoder exchanged for a set-based encoder ~\citep{yang2018foldingnet}. 

\begin{figure}
\includegraphics[width=\textwidth]{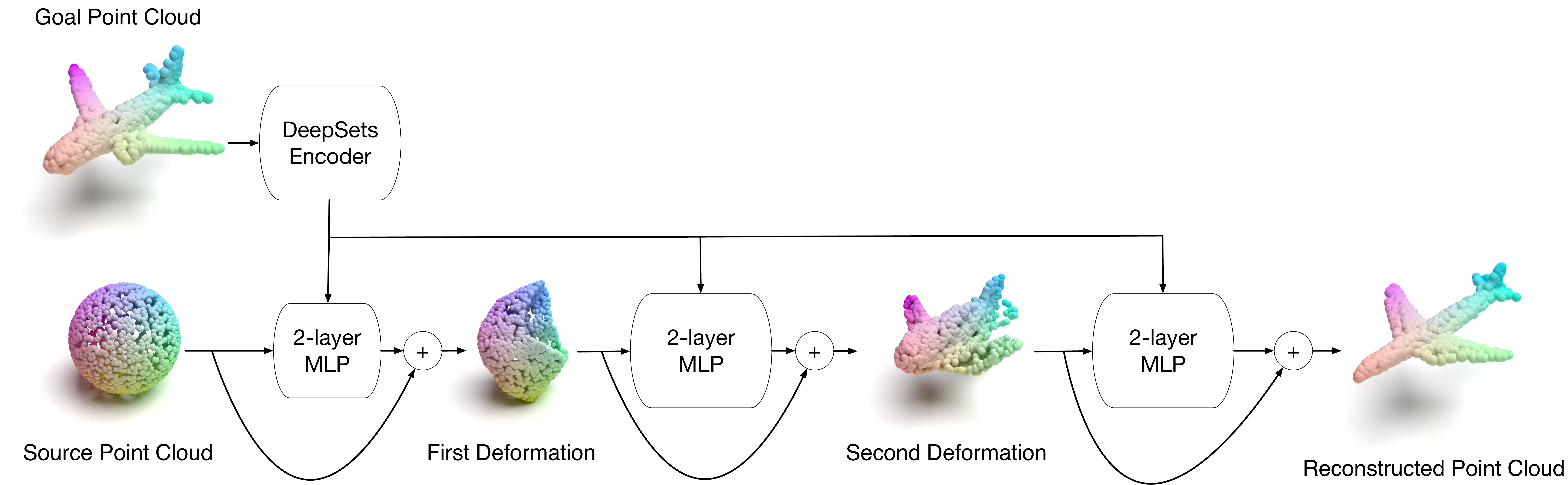}
\caption{A residual structure allows for gradual deformation from a random sphere to the final shape.}
\label{architecture}
\end{figure}

On top of this basic framework, we introduce a forward deformation network (going from a random sphere to the goal point cloud) and a backward pass (from the goal point cloud to a sphere). Training both of these networks simultaneously is meant to regularize the transformation, as inspired by the computer graphics community's requirement for an invertible function without limiting ourselves to models with an analytic inverse. The forward architecture is depicted in Figure \ref{architecture} and the backward architecture is identical.

\subsection{Loss Function}
\label{loss_function_section}

Our loss function encodes our desire to minimize distortion. While the majority of point cloud generative models are trained using Chamfer distance for autoencoder models or maximum likelihood for normalizing flow models, our loss function takes inspiration from the computer graphics community. Note that each $p$ is a point and $\phi(\cdot)$ is our learned function. 

\begin{equation}
    \mathcal{L}_{deform} = \sum_i \sum_{j \in N(i)} (p_i - p_j)^2 - (\phi(p_i)-\phi(p_j))^2
\label{deform_loss}
\end{equation}

Equation \ref{deform_loss} can be seen as an approximation to the Laplacian loss frequently used in mesh generation tasks ~\citep{Wang_2018}. While the true Laplacian would require knowledge of the neighborhood of each point, we can use properties of the sphere to approximate it and then enforce that the neighborhood persists in the output point cloud. For each point on the sphere, its neighborhood may be simply approximated as the $k$-nearest neighbors in Euclidean distance. This allows us to define the neighborhood function $N(\cdot)$ required in equation \ref{deform_loss}. Our final loss function is a weighted combination of the Chamfer loss in both directions and the deformation loss. 

\section{Experimental Results}

\subsection{Dataset}

In order to train and test our model, we sample points uniformly from the surface of meshes provided in the ShapeNet dataset (\cite{shapenet2015}). The 51,300 meshes cover 55 distinct categories including airplanes, cars, lamps, and doors. All of the results in the following sections are trained on a portion of each category. 

\subsection{Metrics and Results}

Taking our cue from previous works, we will use Distance to Face (D2F) and Coverage as measures for the quality of our reconstruction ~\citep{li2018point}. These metrics are used by~\cite{lucic2018gans} to evaluate GANs with the different names called precision and recall.
\begin{equation}
    D 2 F\left(\left\{x_{i}\right\}_{i=1}^{n},\left\{F_{j}\right\}_{j=1}^{m}\right)=\frac{1}{n} \sum_{i=1}^{n} \min _{j} \mathcal{D}\left(x_{i}, F_{j}\right)
\end{equation}

\begin{equation}
    \text{Cov}(\left(\left\{x_{i}\right\}_{i=1}^{n},\left\{F_{j}\right\}_{j=1}^{m}\right) = \frac{|\{ F : \exists x_i, F = \argmin_{F} D(x_i, F) \}|}{m}
\end{equation}

The results in Table \ref{table_results} show that our topologically motivated approach is competitive with traditional methods for point cloud generation. In other words, we do not pay a substantial cost for incorporating topological similarity into our loss function.

\begin{figure}
\begin{subfigure}{.3\textwidth}
  \centering
  \includegraphics[width=.8\linewidth]{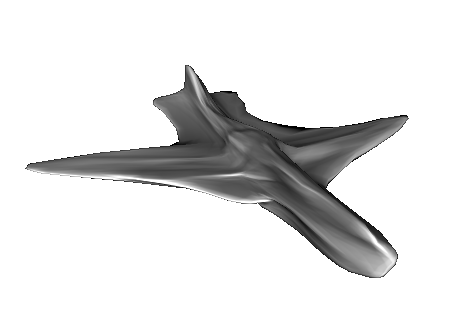}
  \caption{Airplane}
  \label{fig:airplane_mesh}
\end{subfigure}%
\begin{subfigure}{.3\textwidth}
  \centering
  \includegraphics[width=.8\linewidth]{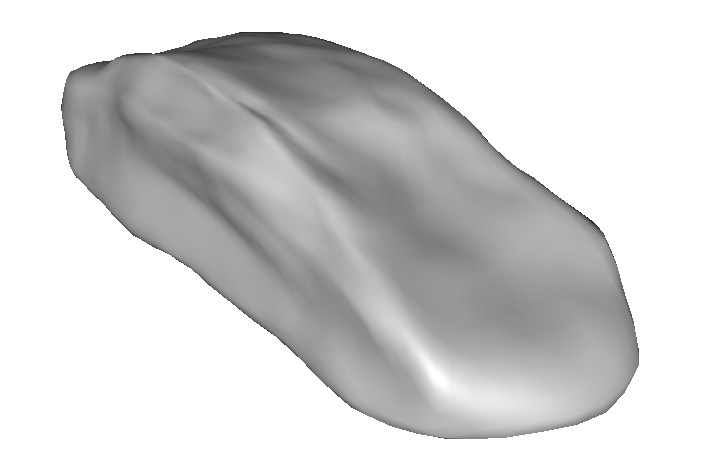}
  \caption{Car}
  \label{fig:car_mesh}
\end{subfigure}%
\begin{subfigure}{.3\textwidth}
  \centering
  \includegraphics[width=.5\linewidth]{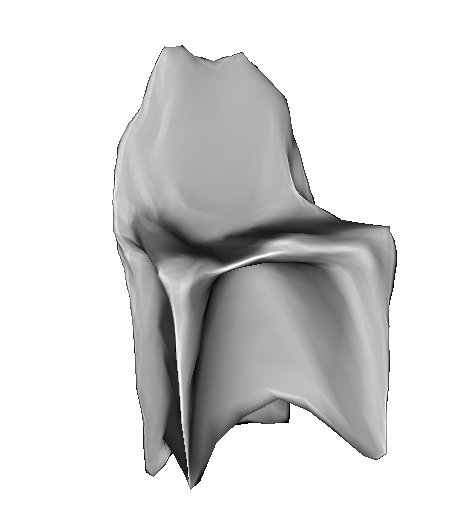}
  \caption{Chair}
  \label{fig:chair_mesh}
\end{subfigure}
\caption{Meshes automatically generated by our approach.}
\label{fig:meshes}
\end{figure}

While these metrics give us an insight into the quality of our reconstruction, their failure to capture structure leads to its poor performance as a measure of accuracy for the underlying surface the point cloud describes. As can be seen in Figure \ref{fig:meshes}, our model is able to produce plausible meshes without a secondary meshing procedure. This is accomplished simply by feeding the vertices of a sphere mesh into our pretrained network. While omitted here for brevity, our ablation experiments show that omitting our deformation loss leads to under-constrained transformations that cause intersecting faces.  

\begin{table}[]
\centering
\begin{tabular}{l|llll|llll}
\toprule
\multirow{2}{*}{Category} & \multicolumn{4}{c|}{D2F ($\times 10^{-2}$)} & \multicolumn{4}{c}{Coverage ($\times 10^{-2}$)} \\
 & \multicolumn{1}{l}{PF} & \multicolumn{1}{l}{FN} & \multicolumn{1}{l}{PCGAN} & \multicolumn{1}{l}{Ours} 
 & \multicolumn{1}{|l}{PF} & \multicolumn{1}{l}{FN} & \multicolumn{1}{l}{PCGAN} & \multicolumn{1}{l}{Ours}\\
\midrule
\multicolumn{1}{l|}{Airplanes} & \multicolumn{1}{l}{2.35} & \multicolumn{1}{l}{1.59} & \multicolumn{1}{l}{\textbf{1.44}} & \multicolumn{1}{l|}{1.59}
& \multicolumn{1}{l}{2.00} & \multicolumn{1}{l}{5.44} & \multicolumn{1}{l}{\textbf{7.40}} & \multicolumn{1}{l}{6.71}\\

\multicolumn{1}{l|}{Benches} & \multicolumn{1}{l}{7.96} & \multicolumn{1}{l}{2.23} & \multicolumn{1}{l}{\textbf{1.92}} & \multicolumn{1}{l|}{2.34}
& \multicolumn{1}{l}{11.8} & \multicolumn{1}{l}{11.5} & \multicolumn{1}{l}{\textbf{13.8}} & \multicolumn{1}{l}{12.66}\\

\multicolumn{1}{l|}{Cars} & \multicolumn{1}{l}{3.67} & \multicolumn{1}{l}{3.98} & \multicolumn{1}{l}{1.46} &  \multicolumn{1}{l|}{\textbf{1.24}}
& \multicolumn{1}{l}{\textbf{1.62}} & \multicolumn{1}{l}{1.01} & \multicolumn{1}{l}{1.44} &  \multicolumn{1}{l}{1.31}\\

\multicolumn{1}{l|}{Chairs} & \multicolumn{1}{l}{8.80} & \multicolumn{1}{l}{\textbf{7.35}} & \multicolumn{1}{l}{7.55} &  \multicolumn{1}{l|}{7.73} 
& \multicolumn{1}{l}{\textbf{17.2}} & \multicolumn{1}{l}{13.2} & \multicolumn{1}{l}{14.0} &  \multicolumn{1}{l}{14.57}\\

\multicolumn{1}{l|}{Cups} & \multicolumn{1}{l}{7.11} & \multicolumn{1}{l}{\textbf{1.97}} & \multicolumn{1}{l}{2.11} &  \multicolumn{1}{l|}{2.21} 
& \multicolumn{1}{l}{\textbf{19.1}} & \multicolumn{1}{l}{12.5} & \multicolumn{1}{l}{15.0} &  \multicolumn{1}{l}{14.73}\\

\multicolumn{1}{l|}{Guitars} & \multicolumn{1}{l}{5.02} & \multicolumn{1}{l}{5.76} & \multicolumn{1}{l}{\textbf{0.995}} &  \multicolumn{1}{l|}{1.29}
& \multicolumn{1}{l}{\textbf{3.80}} & \multicolumn{1}{l}{2.29} & \multicolumn{1}{l}{3.43} &  \multicolumn{1}{l}{3.56} \\

\multicolumn{1}{l|}{Lamps} & \multicolumn{1}{l}{12.9} & \multicolumn{1}{l}{10.7} & \multicolumn{1}{l}{\textbf{2.84}} &  \multicolumn{1}{l|}{3.17}
& \multicolumn{1}{l}{8.94} & \multicolumn{1}{l}{6.85} & \multicolumn{1}{l}{\textbf{12.7}} &  \multicolumn{1}{l}{12.29} \\

\multicolumn{1}{l|}{Laptops} & \multicolumn{1}{l}{8.55} & \multicolumn{1}{l}{4.86} & \multicolumn{1}{l}{\textbf{1.20}} &  \multicolumn{1}{l|}{1.30}
& \multicolumn{1}{l}{9.47} & \multicolumn{1}{l}{8.80} & \multicolumn{1}{l}{\textbf{12.3}} &  \multicolumn{1}{l}{11.66}\\

\multicolumn{1}{l|}{Sofas} & \multicolumn{1}{l}{6.31} & \multicolumn{1}{l}{6.15} & \multicolumn{1}{l}{1.87} & \multicolumn{1}{l|}{\textbf{1.66}}
& \multicolumn{1}{l}{\textbf{11.6}} & \multicolumn{1}{l}{8.11} & \multicolumn{1}{l}{11.1} & \multicolumn{1}{l}{9.95}\\

\multicolumn{1}{l|}{Tables} & \multicolumn{1}{l}{10.9} & \multicolumn{1}{l}{9.88} & \multicolumn{1}{l}{\textbf{9.86}} & \multicolumn{1}{l|}{10.1} 
& \multicolumn{1}{l}{13.1} & \multicolumn{1}{l}{12.0} & \multicolumn{1}{l}{12.3} & \multicolumn{1}{l}{\textbf{13.5}} \\
\bottomrule
\end{tabular}
\vspace{0.5em}
\caption{Our method is comparable with other methods for point cloud reconstruction.}
\label{table_results}
\end{table}

Because our model is limited to deformations between topologically similar objects, the standard reconstruction metrics show a decrease in performance when compared to objects trained for reconstruction loss alone. These shortcomings may be due to the deformation loss incentivizing the deletion of smaller details. 

\section{Conclusions}

Traditionally, generative models for point clouds have been based entirely on the properties of sets: permutation invariance and conditional independence of the points given the underlying shape. Although these properties are crucial for efficiently modeling point cloud distributions, they ignore the relationship between point cloud and mesh, making mesh generation less effective. Our preliminary results show that methods that incorporate this knowledge can be trained solely on point sets and yet produce a generative process for meshes. Our hope is that this progress motivates further research into how set models can benefit from external structure, either as regularization or as a means for improving downstream tasks. 

\clearpage
\bibliographystyle{unsrtnat}
\bibliography{main}

\end{document}